\documentclass{article}
\pdfoutput=1

\usepackage[T1]{fontenc}
\usepackage[utf8]{inputenc} 
\usepackage{times}
\usepackage{graphicx}
\usepackage{url}
\usepackage{alltt}
\usepackage{mdwlist}
\usepackage[leqno]{amsmath}
\usepackage{amsfonts}
\usepackage{amssymb}
\usepackage{algorithmic}
\usepackage{algorithm}
\usepackage{comment}
\usepackage{tabularx}
\usepackage{multirow}

\usepackage[usenames,dvipsnames]{xcolor}

\usepackage{pgf,pgfarrows,pgfnodes} 
\usepackage{tikz} 
\usetikzlibrary{arrows}
\usetikzlibrary{shapes}
\usetikzlibrary{decorations.text}
\usetikzlibrary{trees}
\usetikzlibrary{snakes}
\usetikzlibrary{plotmarks}
\usetikzlibrary{fit}

\newcommand{\smiley}[3]{
\draw (#1,#2) node[shape=circle,minimum size=.8cm,fill=blue!20] (#3) {};
\draw[fill=white,color=white] (#1,#2) ++(-2mm,1mm) circle (.8mm);
\draw[fill=white,color=white] (#1,#2) ++(2mm,1mm) circle (.8mm);
\draw (#1,#2) ++(0mm,1mm) -- ++(0mm,-2mm);
\filldraw[fill=white,color=white] (#1,#2) ++(-1.5mm,-2mm)  arc (225:315:2.2mm) -- cycle;
}

{\small\begin{alltt}{}}{\end{alltt}}

{\scriptsize\begin{alltt}{}}{\end{alltt}}

\title{Evolving knowledge through negotiation\footnote{This is not an abstract of the presentation I gave at the Dagstuhl Seminar 12221 on ``cognitive approaches on the semantic web'' organised by Deirdre Gentner, Pascal Hitzler, Kai-Uwe Kühnberger and Frank van Harmelen, but an account of how the seminar made me change my position.}}
\author{J\'er\^ome Euzenat\\
INRIA \& LIG, Grenoble, France\\
Jerome.Euzenat@inria.fr}

\date{\today}

\begin{document}

\maketitle

\begin{abstract}
Semantic web information is at the extremities of long pipelines held by human beings. They are at the origin of information and they will consume it either explicitly because the information will be delivered to them in a readable way, or implicitly because the computer processes consuming this information will affect them. Computers are particularly capable of dealing with information the way it is provided to them. However, people may assign to the information they provide a narrower meaning than semantic technologies may consider. This is typically what happens when people do not think their assertions as ambiguous. Model theory, used to provide semantics to the information on the semantic web, is particularly apt at preserving ambiguity and delivering it to the other side of the pipeline. Indeed, it preserves as much interpretations as possible. This quality for reasoning efficiency, becomes a deficiency for accurate communication and meaning preservation. Overcoming it may require either interactive feedback or preservation of the source context.
Work from social science and humanities may help solving this particular problem.

\end{abstract}



\section{The problem of semantic interoperability on the semantic web}

Semantic web information is at the extremities of long pipelines held by human beings. They are at the origin of information and they will consume it either explicitly because the information will be delivered to them in a readable way, or implicitly because the computer processes consuming this information will affect them.

Computers are particularly capable of dealing with information the way it is provided to them. However, people may assign to the information they provide a narrower meaning than semantic technologies may consider. This is typically what happens when people do not think that their assertions are ambiguous.

\begin{figure}
\begin{center}

\begin{tikzpicture}

\smiley{0}{0}{emiter};

\draw (5,0) node [fill=yellow, minimum height=7cm,single arrow, draw=none] {message}; 

\smiley{10}{0}{receptor};

\node[cloud callout={(3,1.5),cloud puff arc=120},aspect=2.5,callout pointer segments=6,callout absolute pointer={(emiter.north east)}, fill=blue!50] at (0,1.5) at (2,3.75) {Forest};

\draw (5,0) node[scale=.06] {\pgfimage{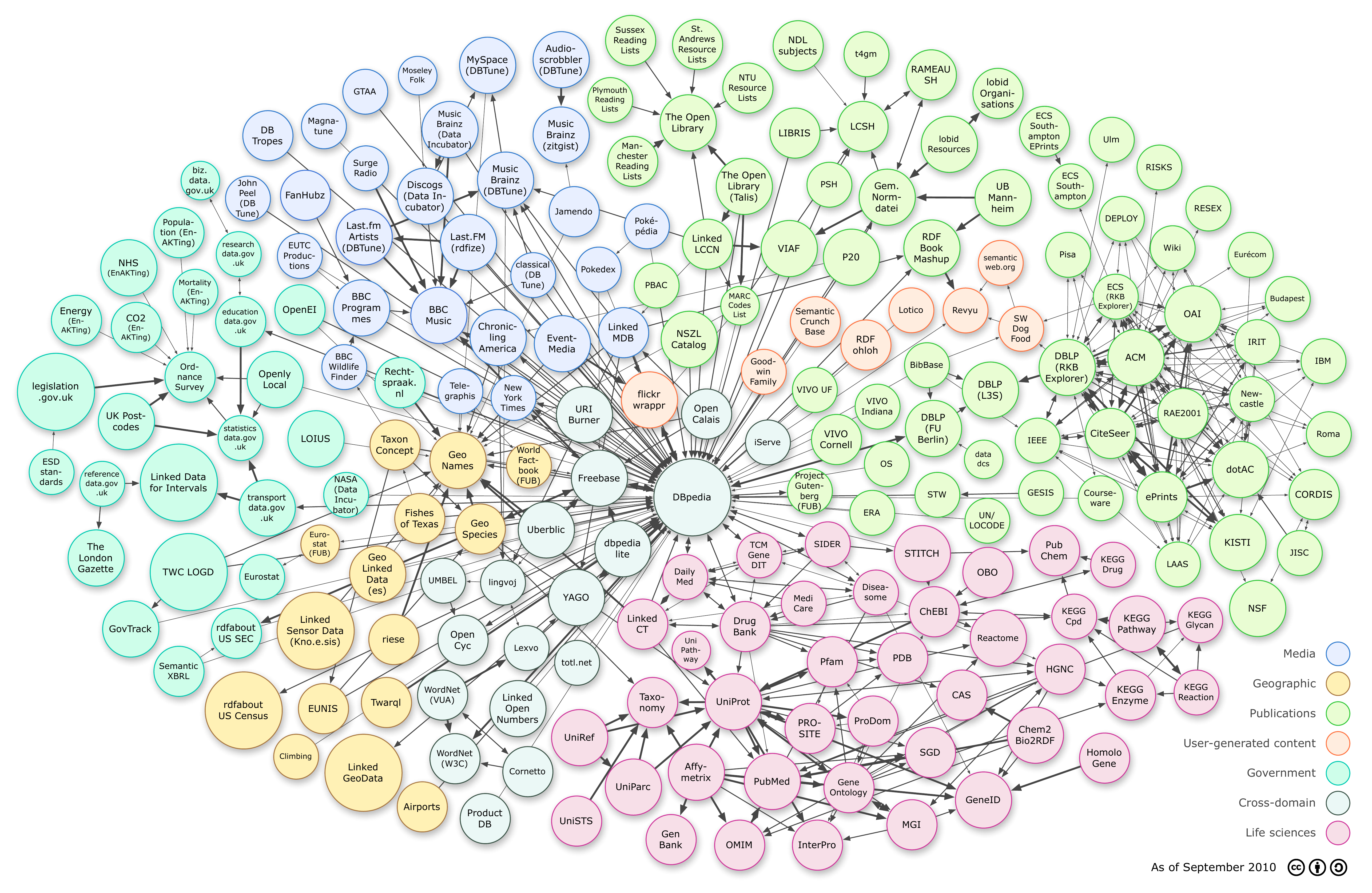}};
\draw (2.5,0) node {$\sqcap\equiv\neg\sqcup$};

\node[cloud callout={(3,1.5),cloud puff arc=120},aspect=2.5,callout pointer segments=6,callout absolute pointer={(receptor.north west)}, fill=red!50] at (0,1.5) at (8,3.75) {Bosquet?};

\draw (5,2) node[cloud, fill=gray!40, text width=1cm,aspect=2.5] {};
\draw (6.5,1.5) node[cloud, fill=gray!40, text width=1cm,aspect=2.5] {};
\draw (3.6,2.7) node[cloud, fill=gray!40, text width=1cm,aspect=2.5] {};
\draw (5.5,3) node[cloud, fill=gray!40, text width=1cm,aspect=2.5] {};
\draw (4.5,5) node[cloud, fill=gray!40, text width=1cm,aspect=2.5] {};
\draw (6.3,4.5) node[cloud, fill=gray!40, text width=1cm,aspect=2.5] {};

\end{tikzpicture}

\end{center}
\caption{A message describing a particular context may be interpreted in various ways, even through model theoretic semantics.}\label{fig:mt}
\end{figure}
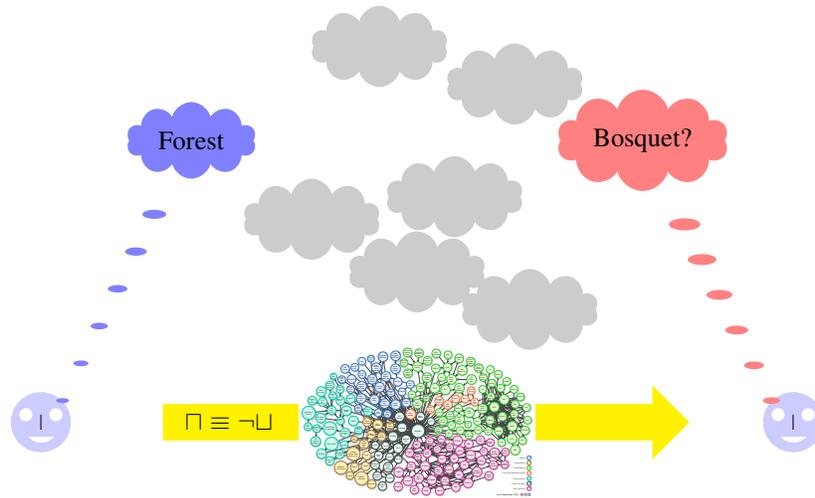

Model theory, used to provide semantics to the information on the semantic web, is particularly apt at preserving ambiguity and delivering it to the other side of the pipeline. Indeed, it preserves as much interpretations as possible. This quality for reasoning efficiency, becomes a deficiency for accurate communication and meaning preservation.

This may be explained by Figure~\ref{fig:mt}: although the first person is thinking having described a forest, his message may have many interpretations. In model theoretic terms, his message enjoys many models (depicted by the little clouds). However, among these models, the receptor may deliberately choose one of these different from the one intended by the emitter.

As far as logic is concerned, this is not really a problem: the computer system is supposed to answer about what is true in all models. Hence, it will answer according to the models chosen by the two people.

But for communication, this is not ideal: the interlocutors are thinking about different things feeling that they understand each others.
Overcoming this may require either interactive feedback or preservation of the source context.

\section{Carrying context}

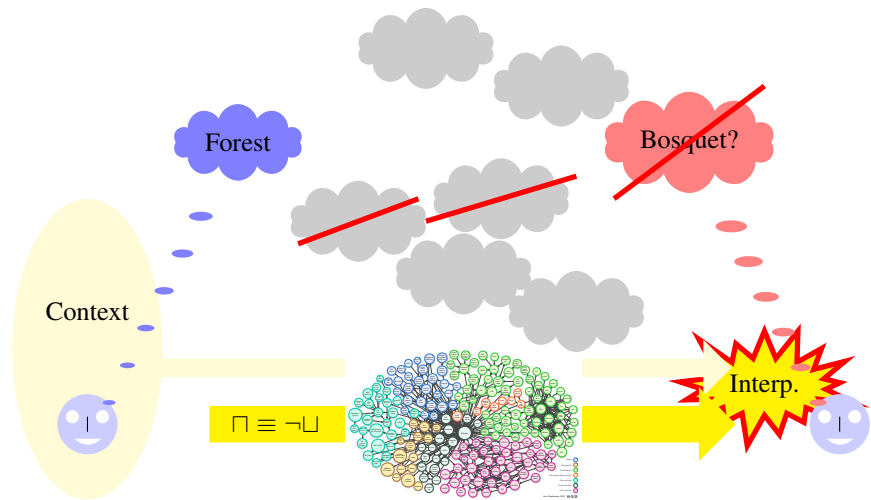
\begin{figure}
\begin{center}

\begin{tikzpicture}

\fill[yellow!20] (0,1) ellipse [x radius=1cm,y radius=2cm];
\draw (0,1.5) node {Context};

\draw (9,.5) node[starburst, fill=yellow,draw=red, line width=2pt] {Interp.};


\draw (4.5,.75) node [fill=yellow!20, minimum height=8cm,single arrow, draw=none] {}; 

\smiley{0}{0}{emiter};

\draw (5,0) node [fill=yellow, minimum height=7cm,single arrow, draw=none] {message}; 

\smiley{10}{0}{receptor};

\node[cloud callout={(3,1.5),cloud puff arc=120},aspect=2.5,callout pointer segments=6,callout absolute pointer={(emiter.north east)}, fill=blue!50] at (0,1.5) at (2,3.75) {Forest};

\draw (5,0) node[scale=.06] {\pgfimage{lod-datasets_2010-09-22_colored}};
\draw (2.5,0) node {$\sqcap\equiv\neg\sqcup$};

\node[cloud callout={(3,1.5),cloud puff arc=120},aspect=2.5,callout pointer segments=6,callout absolute pointer={(receptor.north west)}, fill=red!50] at (0,1.5) at (8,3.75) {Bosquet?};

\draw (5,2) node[cloud, fill=gray!40, text width=1cm,aspect=2.5] {};
\draw (6.5,1.5) node[cloud, fill=gray!40, text width=1cm,aspect=2.5] {};
\draw (3.6,2.7) node[cloud, fill=gray!40, text width=1cm,aspect=2.5] {};
\draw (5.5,3) node[cloud, fill=gray!40, text width=1cm,aspect=2.5] {};
\draw (4.5,5) node[cloud, fill=gray!40, text width=1cm,aspect=2.5] {};
\draw (6.3,4.5) node[cloud, fill=gray!40, text width=1cm,aspect=2.5] {};

\draw[red, line width=2pt] (2.8,2.4) -- (4.4,3);
\draw[red, line width=2pt] (4.5,2.7) -- (6.5,3.3);
\draw[red, line width=2pt] (7,3) -- (9,4.5);

\end{tikzpicture}

\end{center}
\caption{Carring contextual information allows for reducing the interpretation of a message.}\label{fig:contextcarrying}
\end{figure}

One possible solution to this problem is to carry context with the message, illustrated by Figure~\ref{fig:contextcarrying}. 
Because context narrows the meaning of messages, carrying context with them will somewhat freeze or at least better restrict its interpretation.

From a logical standpoint, this is what has to be done: if the speaker is not happy about his message being misunderstood, he always can add axioms, which will make it less ambiguous.

\begin{figure}
\begin{center}

\begin{tikzpicture}[scale=.5]

\fill[yellow!50] (1,1) ellipse [x radius=6.25cm,y radius=3.25cm];
\fill [decorate,decoration={text along path,text=Semantics, text align=fit to path}]
(-.3,3.6) arc (100:80:8cm);

\fill[yellow] (1,1) ellipse [x radius=5cm,y radius=2.5cm];
\fill [decorate,decoration={text along path,text={Semiology, hermeneutics}, text align=fit to path}]
(-2,2.5) arc (110:70:9cm);

\fill[green!50] (1,1) ellipse [x radius=3.5cm,y radius=1.75cm];
\fill [decorate,decoration={text along path,text=Rhetorics, text align=fit to path}]
(-.5,2) arc (105:75:6cm);

\fill[green!80!black] (1,1) ellipse [x radius=2cm,y radius=1cm];
\draw (1,1) node {Pragmatics, context};

\end{tikzpicture}

\end{center}
\caption{Discipline inclusions for considering the context of a message.}\label{fig:discincl}
\end{figure}
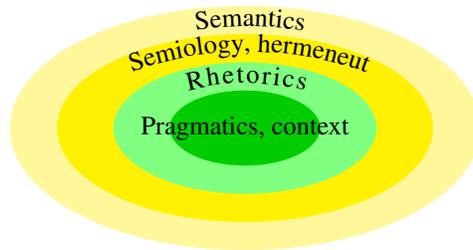

This context may be described by particular structures.
These would allow the interlocutor to know what is the purpose of the message. 
Beside computer science, several types of studies are considering this context and how it acts in interpretation (Figure~\ref{fig:discincl}): 
\begin{description}
\item[Semantics] is the counterpart of model theory for natural language analysis;
\item[Semiology] generalises it to wider sign systems and less grammatical ways to determine sense; \textbf{hermeneutics} is the process of ascribing meaning to a text;
\item[Rhetorics] is concerned with how messages are expressed in order to achieve a particular goal (convincing); this is also related to \textbf{speech act theory};
\item[Pragmatics] is more directly concerned with taking context into account in interpretation.
\end{description}
All these disciplines, from the social sciences and the humanities, could help providing ways to consider context.
Some such structures partially coming cognitive science work are scripts and situations used in language understanding \cite{schank1977a}, communication protocols used in multi agent systems and inspired by speech act theories \cite{searle1985a}, language games used in language science investigations \cite{steels2012a}.

In model theoretic terms, this will act exactly as adding axioms to a theory: adding axioms will restrict the set of models because models are interpretations satisfying all the axioms. This may push the interlocutor to switch to another privileged model.

The problem evolves into determining what part of context should be carried out. The received pieces of context have to be considered together with the context of the receptor. Indeed, ideally it is the smallest piece of knowledge that allows for reconstructing the meaning. However, this is very difficult to determine, even if the interlocutor is known. But what if the receptor is unknown, like in the case of a book or the semantic web? It is difficult to decide which piece of context has to be forwarded with the actual message. It is also not guaranteed that it will be sufficient to interpret the message 1000 years from now, let alone by extra-terrestrial intelligence.

\begin{figure}
\begin{center}
\includegraphics{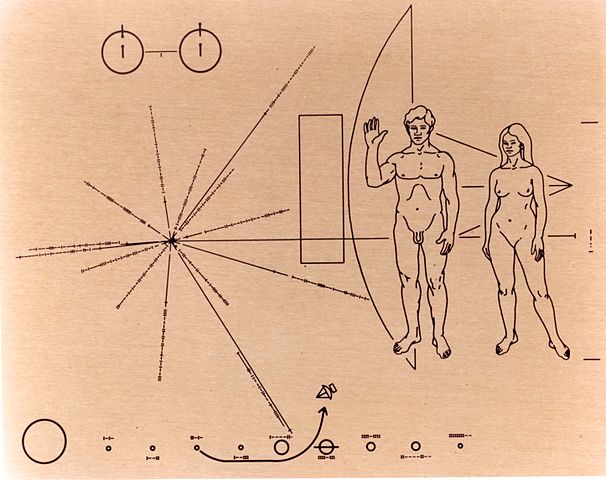}
\end{center}
\caption{ The Pioneer plaque, designed as a message to extraterrestrial beings. The plaque shows where Pioneer came from and who sent it. The barbell at top left represents the hydrogen atom; the radial pattern at left center shows the position of the Earth with respect to pulsing stars. Photo courtesy of NASA.}\label{fig:pionner}
\end{figure}

A good example of this is the Pionner plaque (see Figure~\ref{fig:pionner}) in which the target is any intelligent being! It is particularly unclear that this plaque reaches its target.

Some go even further away from this: in interpretative semantics \cite{duteil2004a}, a text has no sense in itself. Its meaning is only given by the act of reading. If the same person reads the same text several times, then she will interpret it differently each time (at least, for the simple way of the memory of the previous times).

Concluding that texts have no sense is going too far: books are written since thousands of years and they are interpreted quite consistently overall. So, text semantics has some basis. However, interpretative semantics has some either.

\section{Dialogue and feedback}

I already presented these issues 10 years ago \cite{euzenat2000d} and more recently.
Then I had been criticised for having a too simplistic view of communication without any hint about what to do about it.

The discussions at Dagstuhl, especially with Gudrun Ziegler who objected at the slightest form of structuralist commitment, were very fruitful.
For human users, disambiguation is usually obtained through dialog (communication is not a one-way process). 
The receptor provides hints about what part of context is missing to interpret the message, either by direct interaction, or by making interpretation mistakes. 
So communication occurs through a constantly adaptive process which can change all rules: lexicon, grammar, pragmatic.
Writers have to guess what the necessary context is.

This made me think about how to try to compensate the absence of a real dialog with the availability of semantic web technologies and huge data.
In the case of computerised tools like the semantic web, it is however possible to have a dialog through feed-back on how data is useful. 
Certainly an interesting question is how to account for this feed-back to ever improve the communicated data. 
Can interaction or dialog theory developed in social science and humanities be used for that purpose?

This is what should be achieved, at web scale, through feed-back\dots 
When Frank van Harmelen says in his keynote ``Heterogeneity is solved socially'', he is still thinking about human beings mostly interacting directly.
However, the work on emergent semantics and the chinese whispering game introduced by Philippe Cudré-Mauroux \cite{cudre2006a} or the situated alignment method proposed by Manuel Atencia \cite{atencia2010a} are already examples of feed-back mostly provided by programmes. This is somewhat connected to the work of Luc Steels on language evolution \cite{steels2012a} which aims at having agents evolving the language they use for communication through language games.

In principle, this could be achieved by using retroaction to learn the meaning (and the rules that govern its composition). This can be achieved by taking as feedback the structure that are used and those which trigger problems in the current semantic web: ontologies, linked data, alignments. Heterogeneous sources of data are a unique chance of automatically broadening and restricting meaning. Of course, we do not have a full practical set of tools for evolving meaning, but this should be tried at the semantic web scale.

The goal is not to have a fixed meaning for anything, but to use dynamics for adapting to current use.

\section{Conclusion}

Interoperability and understanding on the semantic web is not easier than in social and human context. 
Hence, the hints obtained from social sciences and the humanities may contribute approaching these problem.
Robert Goldstone's synthesis of the Dagstuhl seminar achieved to convince me that we should not try to solve the problem of interoperability in a one-way static structural setting, but in a social dynamic (yet structural) setting.

\bibliographystyle{plain}
\bibliography{communication}

\begin{thebibliography}{1}

\bibitem{atencia2010a}
Manuel Atencia.
\newblock {\em Semantic alignment in the context of agent interaction}.
\newblock PhD thesis, Universitat Autonoma de Barcelona (ES), 2010.

\bibitem{cudre2006a}
Philippe Cudr\'e-Mauroux.
\newblock {\em Emergent Semantics: Rethinking Interoperability for Large Scale
  Decentralized Information Systems}.
\newblock PhD thesis, EPFL, Lausanne (CH), 2006.

\bibitem{duteil2004a}
Carine Duteil-Mougel.
\newblock Introduction à la sémantique interprétative.
\newblock {\em Texto!}, 2004.

\bibitem{euzenat2000d}
{Jérôme} {Euzenat}.
\newblock Towards formal knowledge intelligibility at the semiotic level.
\newblock In {\em Proc. ECAI workshop on applied semiotics: control problems,
  Berlin (DE)}, pages 59--61, 2000.

\bibitem{schank1977a}
Roger Schank and Robert Abelson.
\newblock {\em Scripts, Plans Goals and Understanding}.
\newblock Lawrence Erlbaum, New Jersey, 1977.

\bibitem{searle1985a}
John Searle and Daniel Vanderveken.
\newblock {\em Foundations of Illocutionary Logic}.
\newblock Cambridge University Press, Cambridge (UK), Cambridge, 1985.

\bibitem{steels2012a}
Luc Steels, editor.
\newblock {\em Experiments in cultural language evolution}.
\newblock John Benjamins, Amsterdam (NL), 2012.

\end{thebibliography}

\end{document}